# Multi Defect Detection and Analysis of Electron Microscopy

# Images with Deep Learning


Mingren Shen[1], Guanzhao Li[1], Dongxia Wu[7], Yuhan Liu[3], Jacob R.C. Greaves[1], Wei

Hao[3], Nathaniel J. Krakauer[3], Leah Krudy[4], Jacob Perez[1], Varun Sreenivasan[3], Bryan

Sanchez[5], Oigimer Torres[6], Wei Li[1], Kevin.G. Field[2], Dane Morgan[1]

[1] Department of Materials Science and Engineering, University of Wisconsin-

Madison, Madison, Wisconsin, 53706, USA

[2] Materials Science and Technology Division, Oak Ridge National Laboratory;

Current address: Nuclear Engineering and Radiological Sciences Department,

University of Michigan-Ann Arbor

[3] Department of Computer Sciences, University of Wisconsin–Madison, Madison,

Wisconsin, 53706, USA

[4] Department of Mathematics, Hope College, 141 E 12th Street, Holland Michigan,

49423, USA

[5] Department of Computer Science and Engineering, University of Puerto Rico at

Mayagüez, Mayaguez, 00682, Puerto Rico

[6] Electrical and Computer Engineering Department, University of Puerto Rico at

Mayagüez, Mayaguez, 00681, Puerto Rico




[7] Department of Mathematics, University of Wisconsin-Madison, Madison,

Wisconsin, 53706, USA




**Abstract**

Electron microscopy is widely used to explore defects in crystal structures, but human detecting of defects is often time-consuming, error-prone, and unreliable, and is not scalable to large numbers of images or real-time analysis. In this work, we discuss the application of machine learning approaches to find the location and geometry of different defect clusters in irradiated steels. We show that a deep learning based Faster R-CNN analysis system has a performance comparable to human analysis with relatively small training data sets. This study proves the promising ability to apply deep learning to assist the development of automated analysis microscopy data even when multiple features are present and paves the way for fast, scalable, and reliable analysis systems for massive amounts of modern electron microscopy data.




**INTRODUCTION**

Electron microscopy (EM) is one of the most powerful tools for researchers to extract and collect micrometer down to angstrom scale structural and morphological properties of materials, including repeated structural units (e.g., unit cells of crystals) and defected regions (e.g., grain boundary, impurities, defect clusters). Traditionally, researchers have to manually label defects and repeatedly measure the relevant properties to obtain statistically meaningful values, which is time-consuming, error-prone, inconsistent, and hard to scale[1]. The issue of scaling has become pressing as increasing usage and advancement of EM techniques, such as high-speed detector and automated sample exploration in EM, now generate massive amounts of image data (e.g., up to thousands of images from a single experiment or condition can be generated in minutes) which will keep increasing in the near future[2]. Data on this scale cannot be practically examined by humans, and automated approaches are therefore now necessary to utilize the full power of modern EM. Not surprisingly, the EM community has developed many accurate image data analysis tools that can be effectively deployed to accommodate the large volume of EM data[3,4]. However, due to the complexity of images of material systems, these tools generally still need significant hand-tuning, and in some cases (like counting defects of irradiated materials), human identification of each defect is still the norm.



Well-designed and well-tested automated analysis has proven to be significantly more efficient, repeatable, and standardized than human analysis for discrete cases[2,4,56]. Developing automatic methods for EM image processing has drawn a great deal of interest in the material science community. Automation efforts typically rely on traditional computer vision technology, such as variance hybridized mean local thresholding[7], texture representation, and template matching methods like bag of visual words (BoW)[8,9], key-point matching methods[10,11], Hessian-based Blob boundary detection methods[12] and sometimes obtain better performance[13,14] by utilizing tools from broader areas, such as incorporating synthetic image dataset[15] and using machine learning methods like support vector machine[16] and k-means clustering[2]. However, these efforts typically require extensive human tuning and/or are limited to specific tasks.

Recent developments (<10 years) in deep learning methods[17,18] have demonstrated that object detection in images can be automated with minimal hyperparameters and yield human or even better than human levels of performance. These frameworks are now being adapted towards finding defects in metals, and particularly nuclear materials, including the automated detection of dislocation loops, cavities, precipitates, and line dislocations[5,19–21]. Generally, there are three different approaches for applying deep learning frameworks to defect detection in microscopy images. The first is using the combination of both traditional techniques and deep learning tools[5], for example, Li et al. develops an analysis model that includes a local



visual content descriptor widely used in computer vision called Local Binary Patterns

(LBP)[22] descriptor, feature selecting methods called AdaBoost[23], and Convolutional

Neural Network (CNN) module to screen candidate bounding boxes to obtain the best

performance. This approach is more like an intermediate stage of applying deep learning

since it does not follow the complete end-to-end pattern of deep learning practice[24] but it

helped show that value of new deep learning methods for this class of problems. The

second approach relies on the encoder-decoder framework to find features[25] in EM

images. Examples include using weakly supervised learning methods of encoder-decoder

to study the local atom movements[26], U-Net to study nanoparticle segmentations[27], and a

modified U-Net framework to segment defects in STEM images of steels[19]. The encoder-

decoder framework can extract the most relevant information in images and use the

extracted inner state to do other tasks, but the performance of encoder-decoder

framework relies on the extracted inner state and good performance requires careful

training[25]. The third category of methods are using mature object-detection frameworks,

for example, Chen et al. used Mask R-CNN to study the microstructural segmentation of

aluminum alloy[28] and Anderson, et al. used Faster R-CNN to study helium bubbles in

irradiated X-750 alloy[20]. Here we explore the first use of similar mature object-detection

deep learning methods as this third category, specifically Faster R-CNN, to obtain

properties of dislocation loops with varying Burgers vector and habit plane in neutron-



irradiation related iron-chromium-aluminum (FeCrAl) materials, These materials are important for the development of next generation nuclear reactors[29–31].

Analyzing the locations and sizes of defects in materials that have undergone irradiation is a widely used application of electron microscopy. In such studies, the key properties are the total number and distribution of each type of defect. Typical defects of interest include grain boundaries, precipitates, dislocation lines, dislocation loops, stacking fault tetrahedra, cavities (voids, bubbles), and co-called "black-spot" defects, which are small defect clusters of interstitials and sometimes vacancies[1,32]. For this study, we focus on the dislocation loops formed within a ferritic alloy, where the loops exist on specific habit planes that manifest themselves with different morphologies due to the projection of a 3D volume imaged using EM[33]. Typical microstructural images of irradiated ferritic steels contain four prominent types of defects: (1) open ellipse loops (single ring edge), (2) open ellipse loops (double ring edges), (3) closed solid elliptical loops, (4) closed circular solid dots[33]. Figure 1 shows a sample STEM image containing all four morphologies of loops obtained from a ferritic alloy irradiated in a materials test reactor. In this paper, we used a modern deep learning-based object detection model called Faster Regional CNN (Faster R-CNN)[34], a widely used deep learning based object detection model[17]. We use the Faster R-CNN to develop an automatic defect detection system for all four morphologies commonly observed in irradiated steels with a body-centered cubic structure and then additional post-processing to analyze their geometrical



information (specifically, size and areal density). This paper serves to demonstrate the power of deep learning-based computer vision models for material image studies and suggests the possibility that most aspects of defect analysis may soon be practically automated, and many, if not all, handcrafted feature-based methods may be replaced by deep learning methods.

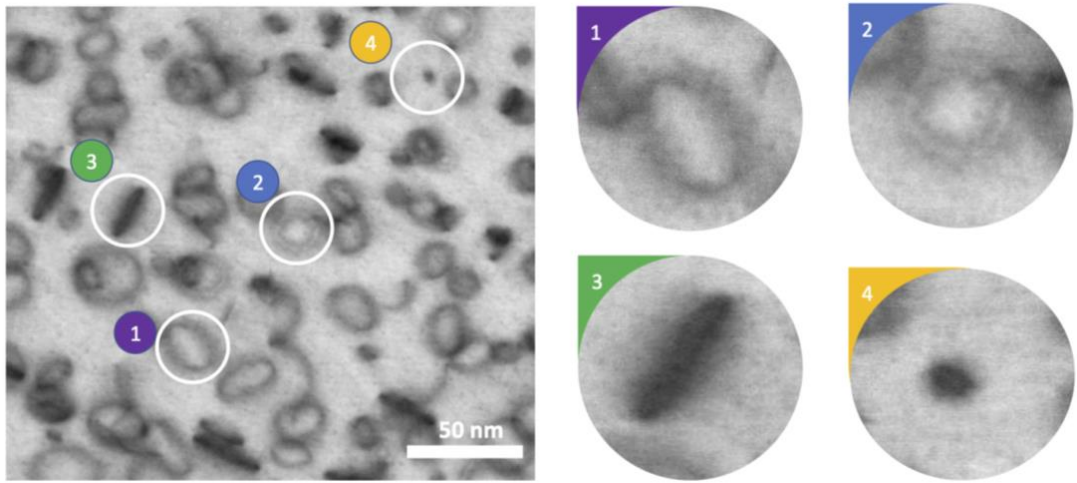

Figure 1. Selected bright field scanning transmission electron microscopy (STEM) image of an irradiated ferritic alloy showing four common morphologies of dislocation loops: (1) open ellipse loops (single ring edge), (2) open ellipse loops (double ring edges), (3) closed elliptical solid loops, (4) closed circular solid dots. Open single edge ellipse loops (1) are dislocation loops with a Burgers vector of $a_0/2\langle 111 \rangle$. Open double edge ellipse loops (2) and closed elliptical solid loops (3) are dislocation loops with a Burgers vector of $a_0\langle 100 \rangle$. Closed circular solid dots (4) are black dot defects with a Burgers vector of either $a_0/2\langle 111 \rangle$ or $a_0\langle 100 \rangle$. Image size: Primary image is 290 × 290 nm; inset scales arbitrary.



Faster R-CNN is a CNN based end-to-end deep learning object detection model that outputs both the object position and its class[34]. As shown in Figure 2, Faster R-CNN is a two-stage detector where the region proposal network (RPN) proposes Region of Interest (ROI), and the following ROI regressor and classifier will fine tune the final output results including the size and position of the object contained bounding boxes and the corresponding object label[34]. Given an image, the shared convolutional layers will extract a feature map from the input image by performing a series of convolution and max pooling operations. Then based on the extracted feature map, the RPN will put a set of predefined anchor boxes on the feature map and output the probability of whether the anchor box belongs to an object of interest or plain background. It worth mentioning that RPN ignores the specific object class of each bounding box and the following ROI regressor and classifier are responsible for the specific class and refined location of the objects. The refining network predicts certain object labels and refines the size and position of each bounding box based on the feature map generated by the ROI-pooling layers[35]. The RPN and ROI components are trained jointly to minimize the loss function sums from both of them[34]. After the Faster R-CNN module A, those images with detected defects are sent to module B to extract geometric information such as defect diameters, as shown in Figure 2.



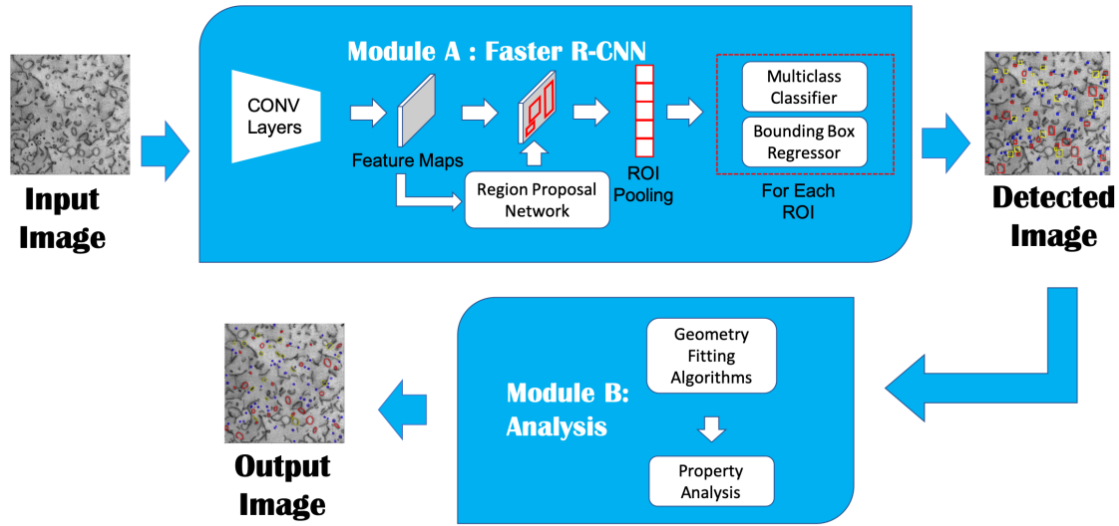

Figure 2. Schematic flow chart of proposed deep learning based automated detection approach. Input micrograph images go through the pipeline of Module A—Faster R-CNN Detector, Module B—Image Property Analysis. After Module A, the loop locations and bounding boxes are identified and then for each identified bounding box, geometry fitting algorithms are called to determine the defect shape and size in Module B.

For details on the data sets, training approaches, and methods of defect identification and analysis, please see Methods section. Please see Data section for summary of all data made available.

**RESULTS**

To assess the machine predictions, four types of approaches were taken. The first approach was a qualitative comparison of machine to human labeled images, where we looked for large fractions of errors, e.g., more than 40%, and for trends in errors that



might indicate a major issue but made no attempt to quantify agreement. This assessment

tests all aspects of the model as it compares to the ground truth human results, which

include the bounding box predictions (the defect detection part of Module A in Figure 2),

the defect type identifications (the categorization part of Module A in Figure 2), and the

geometric shape determination (Module B in Figure 2). The second assessment approach

was a quantitative assessment of the ability to identify a defect, regardless of defect types.

This assessment tested the defect detection part of Module A (see Figure 2). This

assessment was a binary categorization problem and success was quantified with

precision, recall, and F1 score. The third assessment was a quantitative assessment of the

ability to identify a defect type once a defect had been correctly identified and tested the

categorization part of Module A (see Figure 2).  This assessment was a three-category

categorization problem and was quantified using the confusion matrix with precision,

recall and F1 calculated for each class. Finally, the fourth assessment was a quantitative

assessment of the ability to quantify the geometric properties of defects. This assessment

tested the geometric analysis of Module B (see Figure 2) and compared machine and

human predictions of average and standard deviations in size and areal density for each

defect type. We discuss each of the four assessments below and label them assessment 1-

4 for clarity. In all cases the comparisons are made on the test data set described in

Methods section.



**Assessment 1.** After feeding the images into the Faster R-CNN detectors, the resulting detections were plotted on the original images. As shown in Figure 3, the red circles represent the dislocation loops with a Burgers vector of $a_0/2\langle 111\rangle$ (Type 1 in Figure 1), while the yellow and blue circles represent $a_0\langle 100\rangle$ direction loops (Type 2 and 3 in Figure 1) and "black dot" defects (Type 4 in Figure 1) respectively. The data from both human-labeled and machine detected results are plotted in the same manner. More comparisons can be found in Supplement Information Section 1. To a human observer the machine results show strong correlation of bounding box location, defect type identification, and defect shape with the ground truth human labeling which indicates the effectiveness of the proposed automatic defect detection system.

**Assessment 2**. The performance of the detection part of Module A (see Figure 2) of the trained model was evaluated in terms of precision, recall, and F1 score by comparing the detected result with the human labeled result of the 12-image testing set, as shown in Figure 4. The precision describes the percentage of all machine predicted bounding boxes that are judged to have correct positions, and the recall value describes the percentage of all human labeled defects that are identified as in a bounding box by the machine algorithm. F1 is the harmonic mean of the precision and recall which can be used to assess the overall performance of the defect location task[35]. The IoU (Intersection over Union) method was used to determine if a given defect was identified by a bounding box and is described within the provided Methods section. The cutoff IoU, which must be



exceeded to consider the bounding box to have identified the defect, is a hyperparameter that can be fine-tuned based on the purpose of the object detection task[35]. Figure 4 showed a drop in performance as the cutoff IoU increased. This trend agreed with expectations as the higher cutoff IoU meant it was harder for the predicted bounding box to be judged successful. However, setting the cutoff IoU to an extremely small threshold could lead to the problem that the predicted bounding boxes are associated with defects for which only a small part of the defect is actually in the bounding box, which will likely cause problems in the defect identification (Module B) of our model. As a compromise, for all the further assessments in this paper, we used cutoff IoU = 0.4 to determine when the machine predictions were considered to match a given defect. This choice kept nearly optimal performance of the detector (based on Figure 4) and an adequately demanding standard for predictions.



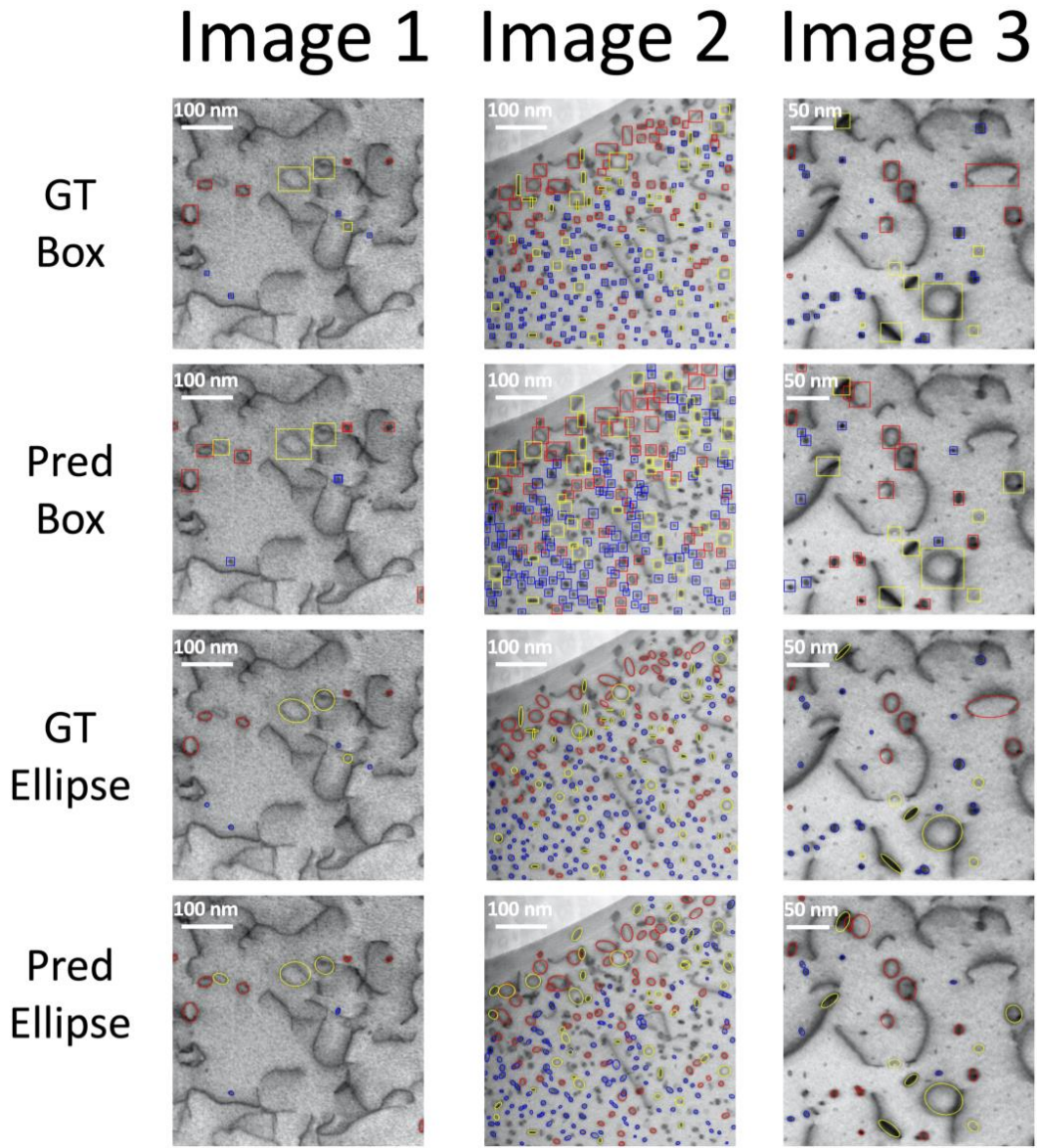

Figure 3.Selected data images to show the detector performance and the fitted ellipse of our automatic

analysis system. These three test images are selected from the test dataset of 12 images (see Methods).  The "Ground

Truth (GT)" shows the bounding box and ellipse human labeling (colored by defect type), the "Prediction (Pred) Box"



shows the predicted bounding boxes (colored by defect type), and the "Prediction (Pred) Ellipse" shows the resulting

fits to the specific defect geometry (colored by defect type as described in the text).

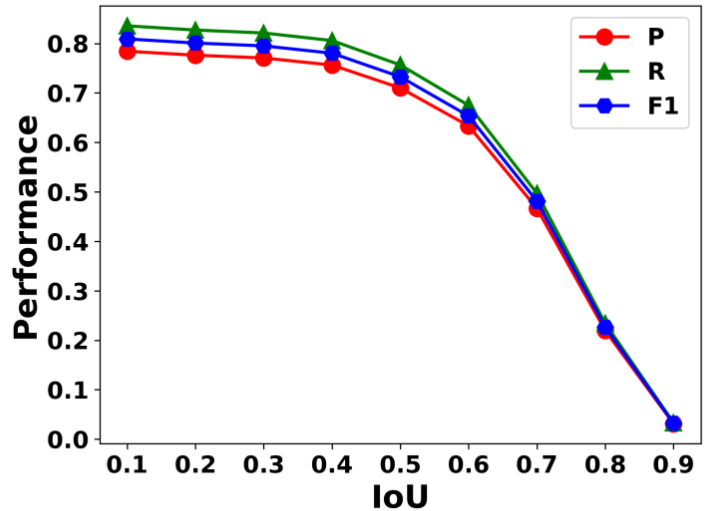

Figure 4. Summary of defect location recognition performance of all types of defects evaluated using precision and

recall metrics, regardless of defect types. The test set contained 12 images, and, for all IoU (Intersection over Union)

values and we used a threshold confidence score 0.25 for Faster R-CNN output. (see Method Section).

**Assessment 3.** Table 1 shows the confusion matrix of the predictions made by

Faster R-CNN detector evaluating its capability to correctly categorize defects. Each row

in the confusion matrix represents a class that is predicted by the detector, and each

column represents a class labeled by human researchers. The diagonal elements of the

table represent the correct classification made by the detector and off diagonal elements



represent errors of different types. We also show the percentage accuracy of each type of

defect in parentheses. The 76%, 87%, and 94% accuracy indicates that once the Faster R-

CNN model locates the defect, it can classify the type of defect based on their

morphology within the image with good accuracy, although some improvement of the

76% value is likely possible for the $a_0\langle 100 \rangle$ loops. We also report the classification

performance using precision, recall and F1 score in Table 2. Given inherent errors of

human performance we take scores for precision, recall and F1 of 0.78 as approximately

the upper limit that can be obtained with the present labeling. Table 2 shows F1 from

about 0.65 to 0.78, which demonstrates significant capabilities but is likely less than can

be achieved, suggesting opportunities for further improvements.

Table 1. Summary of the classification performance for each type of defects at cutoff IoU 0.4. Values in parenthesis

give the % each number represents of the total number of defects in that class as determined by the human labeling.

|  | $a_0/2\langle 111 \rangle$ Loop | Black Dot | $a_0\langle 100 \rangle$ Loop |
|---|---|---|---|
| **a/2⟨111⟩ Loop** | 239 (87.2%) | 21 | 14 |
| **Black Dot** | 17 | 416 (94.3%) | 8 |
| **a⟨100⟩ Loop** | 33 | 13 | 166 (78.3%) |

Table 2. The performance report for each class.

|  | $a_0/2\langle 111 \rangle$ Loop | Black Dot | $a_0\langle 100 \rangle$ Loop |
|---|---|---|---|
| **Precision** | 0.73 | 0.65 | 0.62 |
| **Recall** | 0.83 | 0.71 | 0.72 |
| **F1** | 0.78 | 0.68 | 0.67 |



**Assessment 4.** The second Module provides geometric in formation for each defect through fitting ellipses. While the fits can provide a range of detailed information, we are particularly interested in the arithmetic mean and the associated standard deviation of the defect diameter as well as the areal density in an image for each type of defect. These values are commonly quoted values in literature within irradiated materials studies. Table 3 compares the human labeled arithmetic mean diameters and areal densities to the ones predicted by the automatic analysis system. The discrepancy of arithmetic mean diameter between the human labeled ground truth and predictions is within 10% in all cases, which is considerably less than might be expected for variation among different humans[2] and we consider a strong success. Furthermore, the errors in arithmetic mean diameters are in the range 0.7-1.1 nm, which corresponds to a range of two to nine pixels (based on the range 0.14nm/pixel to 0.87nm/pixel for our test data, see SI Section 1). The errors of about 1 nm correspond to about 5-10% for our data which is somewhat larger than might be expected from direct labeling errors on 10-15nm. Thus, it is unlikely that any human labeling is meaningfully accurate to much below this level. However, the human and machine learning black dot radii do not fall within a 95% confidence interval, suggesting that the algorithm does not yield exactly the same means as the human ground truth. Some errors will come from the machine detection (failures in precision and recall, see Figure 4) and defect type assignment (see Table 1). Additional errors are associated



with intrinsic errors in the machine and human ellipse labeling, where both have some uncertainty due to ambiguity or variances in the morphology of defects in images. In particular, some defects are not well fit by an ellipse (e.g. some have a more rectangular shape, as can be seen in Figure 3), making this form of labeling difficult for both human and machine. Another error to consider is that as the number of pixels per feature goes down, the intrinsic error due to the resolution (pixel/nm) will artificially go up. For instance, a 100 nm loop where the resolution is 1 pixel/nm where the labeling is off by 1 pixel will yield a 1% error. If the labeling is off by 1 pixel for a 5 nm loop, the error will be 20% even though the per pixel error is the same. Seeing as the black dots are all of small arithmetic mean diameter (<10 nm), they will intrinsically have a higher error compared to the other classes where the diameters are 2-3 times larger.

Table 3. Comparison of arithmetic mean defect diameter and standard deviation of mean loop diameter between ground truth labeling and our automatic analysis model prediction with an IoU of 0.4. The values in parenthesis are the relative percentage error between ground truth human labelling results and the automatic analysis results.

| Defect Type | Ground Truth | | | Automatic Analysis Model | | |
|---|---|---|---|---|---|---|
| | Arithmetic Mean diameter (nm) | Standard Deviation of Mean Diameter (nm) | Areal density (m$^{-2}$) | Arithmetic Mean diameter (nm) | Standard Deviation of Mean Diameter (nm) | Areal Density (m$^{-2}$) |
| $a_0/2$ $\langle 111 \rangle$ Loop | 22.4 | 0.7 | $1.77 \times 10^{14}$ | 23.1 (3.1%) | 0.8 | $2.21 \times 10^{14}$ (24.9%) |
| Black Dot | 8.2 | 0.1 | $3.41 \times 10^{14}$ | 9.1 (10.9%) | 0.2 | $4.98 \times 10^{14}$ (46.0%) |



| **a$_0$⟨100⟩ Loop** | 20.3 | 0.8 | $1.32\times10^{14}$ | 22.4 (10.3%) | 0.9 | $1.79\times10^{14}$ (35.6%) |
| --- | --- | --- | --- | --- | --- | --- |

## DISCUSSION

The above results demonstrate that the trained model potentially performs well enough to replace human in a workflow on similar types of data. The precision and recall values for assessing detection in the range 62-83% which are comparable or less than human variation[2] from previous assessments. The machine defect type misidentifications are at the level of 10-25% (see Table 1), and a significant fraction of this variation may also be due to ground truth ambiguities or errors. The final machine predicted diamaters are within a nanometer, approximately 2 pixels in images, which is a level of error that is considered negligible in terms of impact on material properties. To further clarify that the error is negligible for our defect population we have done a sensitivity analysis based on previous studies of hardening from loops. As discussed in Field et al. [31] simple dispersed barrier hardening models suggest that the hardening under irradiation from loops is of the form $\Delta\sigma_y = A\sqrt{d}$ where A is a constant and $d$ is the diameter of the defect. Now consider an error in diameter $d$ defined as $\varepsilon$. The fractional error in $\Delta\sigma_y$ due to the error $\varepsilon$ is $\left(\Delta\sigma_y(d+\varepsilon) - \Delta\sigma_y(d)\right)/\Delta\sigma_y(d) \approx \varepsilon/(2d)$, where the approximate equality holds for $\epsilon \ll d$. For $\varepsilon = 1.7$ nm (which is 2 pixels for our largest pixel sizes, see below) and $d$ = 21.4 nm (our average sizes of a/2⟨111⟩ and a⟨100⟩ defects), we get the fractional error in $\Delta\sigma_y$ as 1.7 nm / (2 * 21.4 nm) ≈ 0.04, which is well within the uncertainty of such



microstructure-based analysis. However, for smaller defects this percentage error could clearly become larger. The errors of diameter between ML results and human results appear to be approximately symmetrically distributed in positive and negative directions and independent of defect density, as shown in detail in SI section 4 and 5. Furthermore, previous studies indicate that the differences of arithmetic mean diameter between different human labelers can be comparable or larger than values found here between the ML and human results[5]. The discrepancy in areal densities is somewhat larger than might be intuitively expected just from the percentage error in the arithmetic mean diameters. However, additional errors are introduced by the exact definition of areal density (see Methods section) and the additional errors introduced by the imperfect precision and recall.

While the exact performance of the present automated approach compared to different human researchers is difficult to determine rigorously there is no doubt that the present approach is much more consistent. Previous studies have shown that different labelers tend to label defects in different ways and even the same person may label the same defects differently even after a short break[2,5]. Such issues can make any given data analysis somewhat unreliable and make it difficult to integrate results across different teams and or time periods in larger analysis efforts. However, once a machine learning model is properly trained, it will yield a unique and reproducible labeling for every image. If the community could converge on a single or small number of models this



could greatly increase the reproducibility in labeling of STEM experiments. That said, models trained on different data and/or different human labeling could give different predictions, so establishing community accepted models is an important part of using these approaches to obtain more consistent results.

The approach applied here is readily scalable to very large data sets. Analyzing a single image with our model on a reasonable state of the art GPU (NVIDIA's GeForce GTX 1080 GPU) takes about 0.1s, so analyzing all the images in a typical experiment can be done easily in minutes, even less if multiple GPUs are used and as GPU and related processors (e.g., TPU) continue to get faster.  As large scale distributed cloud service provider like Google, Amazon and Microsoft are  providing cloud service for deep learning applications with GPU machines[36], it would be easy to scale to process even larger amount of data. Furthermore, significant speedup can likely be obtained if desired. We developed the system with the Python code language and the ChainerCV deep learning framework, both of which were chosen for ease of development not for the high-performance in deployment. Replacing Python with C/C++ or using high-performance deep learning frameworks, e.g. Caffe[37], could potentially accelerate the prediction speed of the current model. In particular, the deep learning community is actively designing new methods to accelerate the running speed of model e.g. model compression, weight sharing, or parameter pruning[38] which could also boost the speed of ours. As an example of how fast deep learning AI algorithms can be, researchers from Google have recently



applied deep learning models for cancer diagnosis on data during the actual process of conducting an optical microscopy experiment[39].

The approach applied here is also readily adapted to new defect types and systems. The present model was trained with only a relatively small amount of training data due to the use of transfer learning[40]. With only modest additional data sets (e.g., on the scale of thousands of defects or possibly fewer) and a few rounds of further training as described in Section II of SI, researchers could likely extend the present model to more defects (e.g. separating the two orientations of 111 loops or adding voids, preexisting dislocations, etc.), different imaging conditions (e.g., changes in microscopes, imaging modes, orientation, focus, etc.), and different materials (e.g. other metal alloys).

There are several areas where significant improvements may be obtainable. The first is that the use of real-world data in the study has led to significant time spent labeling and introducing unavoidable human biases and errors into the deep learning model and its assessment. However, it is possible that simulated images could be both more accurately labeled and generated in large volume, potentially allowing much more accurate models to be trained.

The second area where significant improvement is likely is that deep learning methods for object detection continue to evolve rapidly. In particular, deep learning segmentation models[17], which learn a label for every pixel, could be equally or more accurate and remove the step of fitting contours in a bounding box to get geometric



information. Such an approach was applied recently to automatically detect information about dislocation lines, precipitates and voids in STEM images[19].

**Conclusion**

This study demonstrated a practical deep learning based automatic STEM image defect detection system implemented by incorporating Faster R-CNN for detection and watershed flood algorithm for geometry fitting. Compared with other models proposed before, our model reduced the training effort by utilizing only one module for detection and expanded capability to simultaneously recognize multiple classes of defects. The approach developed here achieved reasonably reliable performance, with an F1 score of 0.78, and predicted sizes and areal densities within the uncertainty of results from human researchers. The automated analysis on NVIDIA's GeForce GTX 1080 GPU processor is about 0.1 s/image, hundreds of times faster than human analysis ($\geq$1 minute/image), and trivially parallelizable and scalable on more processors. The model can also be readily extended to new defects, systems, and conditions with modest training requirements. Thus, our approach provides an accurate, efficient, reproducible, scalable, and extensible method which could replace or greatly enhance human analysis in future studies related to STEM images.



We believe that this framework can be used on many defect and other STEM features simultaneously, eventually providing a general tool for automated analysis across many STEM applications.

## METHODS

### *Data Set Collection*

Data set collection was completed as part of a large-scale effort to characterize iron-chromium-aluminum (FeCrAl) materials neutron-irradiated within the High Flux Isotope Reactor at Oak Ridge National Laboratory. The dataset comprises a series of published[29,31,41] and unpublished data. The data collection was completed over 3 years and spaned a range of different FeCrAl alloys, including model, commercial, and engineering-grade alloys irradiated to light water reactor–relevant conditions (e.g., <15 displacements per atom and temperatures of nominally 285–320°C). Images generation are described in more details in Li et al[5].

### *Data Set Preparation*

We used ImageJ[42,43], an open-source software for analysis of scientific images, to manually label all the training and testing data set. And since STEM images are gray scale and ChainerCV[44] expects input images with RGB channels, some modifications are necessary. We use the direct STEM image gray scale for the R channel. Then we use



modifications of the original image gray scale for the G and B channels. Specifically, following Li et al.[5], for G we use a local contrast enhancement of the original gray scale channel saturated to maximum/minimum and for B we use a Gaussian bluer filter of the original gray scale STEM images. For the local contrast enhancement in channel G, we use the Contrast Limited Adaptive Histogram Equalization (CLAHE), a common algorithm used for local contrast enhancement that makes local detail of STEM image enhanced even in regions that are darker or lighter than most of the image. The Gaussian filter used in channel B represents cases where there might be noises or blurring in the STEM images. The parameters used for CLAHE[45] and Gaussian blur[46] are all from the default parameter setting of scikit-images and details can be found in the references given here for these methods. The purpose of adding two more channels in this way is to improve the model performance and make the model more robust by providing more information about various contrast levels or blurring.

For the training and testing on the Faster R-CNN model, a total of 165 STEM images of irradiated ferritic alloys were collected and labeled. The images were taken at different experimental conditions of temperature and irradiation damage level so that the data includes varying defect sizes, shapes, and areal density. We constructed the ground truth labeling by giving each image in the dataset to at least two groups of at least two researchers per group who together labeled each image in that dataset. In some cases, no absolute consensus could be reached on whether a feature was a defect and/or what type



it had, in which case a best effort was made based on group discussion. Details of the protocol are in the SI.

The test dataset was randomly selected from the complete image dataset, so that the training and test were split by approximately 10:1 ratio. The training dataset was then augmented to 918 images in total, which could provide more training instances without spending more manpower on labeling. The data is augmented by rotating and/or flipping each image in the training set, a standard method previously well established to improved results in some cases[47].

### *Model Training*

The Faster R-CNN model used VGG-16 as its backbone architecture and we adopted the module provided by ChainerCV[44] as the Module  A in Figure 2 and using watershed function provided by OpenCV[48] as the second module. The initial weights of Faster R-CNN was loaded from the pre-trained weights from ImageNet which is a common practice in the deep learning training strategy[40] called transfer learning. Although ImageNet is trained for image classification, not object detection, there are enough similarities in key features to support effective transfer learning of weights. Transfer learning can reduce the amount of data and training time required for good performance[40]. The Faster R-CNN module was optimized with Stochastic Gradient Descent (SGD) on a single Nvidia GeForce GTX 1080 GPU. The best hyper parameter



set was found by performing hyperparameter search of learning rate from $10^{-3}$ to $10^{-6}$ and we adjust the needed iteration numbers correspondingly. The best choice of hyper parameter is a decayed learning rate starting from $10^{-4}$ and each 20000 iterations the learning rate will decay to one tenth of the previous one. In total 90000 iterations were performed, and a learning loss curve is shown in Figure 5. The geometry extraction module needed no training.

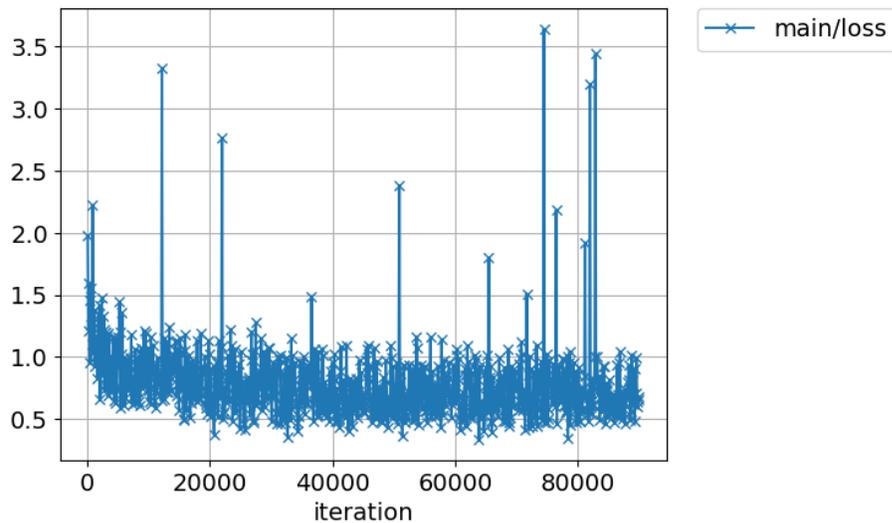

Figure 5. A typical loss curve for Faster R-CNN training.

### *Model Testing*

After the Faster R-CNN module was trained, there were still two important hyperparameter associated with accuracy analysis: the threshold IoU value and the confidence score. IoU stands for Intersection over Union and is an evaluation metric



used to measure the performance of object detection models[17]. IoU is calculated from the ratio of overlap area of a ground truth bounding box and a predicted bounding box to the area of union of two bounding boxes. The range of IoU is from 0 to 1 where 0 means no overlap found between two bounding boxes and 1 means the two bounding boxes are perfectly overlapping. The threshold IoU is the value used to judge the prediction quality of the overlapping of ground truth bounding boxes and prediction bounding boxes. A higher threshold IoU requires more accurate location prediction of the bounding box detector, which will generally reduce performance, but lower the threshold IoU could lead a predicted bounding box to being assigned to no defect or the wrong defect. And another important hyperparameter is the threshold confidence score, a value from 0.0 to 1.0 used by Faster R-CNN internally to discard low confidence proposals in the RPN, and it can change the total number of outputs of Faster R-CNN. We used grid search of the threshold IoU and confidence score to search the best choice of these two values based on maximizing the F1 scores, with confidence score from the list [0.001, 0.005, 0.01, 0.05, 0.1, 0.15, 0.2, 0.25, 0.3, 0.35, 0.4, 0.45, 0.5, 0.55, 0.6] and the threshold IoU from the list [0.1, 0.2, 0.3, 0.4, 0.5, 0.6, 0.7, 0.8, 0.9]. We selected 0.25 as the confidence score for Faster R-CNN and showed the performance changes with 0.4 threshold IoU in Figure 4.

*Geometry Fitting of Analysis Module*



After the Faster R-CNN module was performed on specific image, the analysis module was called to obtain shape and size of the defect contained in bounding box. As shown in the third column in Figure 3, the approach fits the defect with elliptical contours to estimate their actual shapes and diameters. The approach uses the watershed algorithm to identify the pixels that make up the defect contour and then fit those to an ellipse. The watershed algorithm is a widely used technique for image segmentation purposes that views any gray scale image as a topographic surface where the high (e.g. white) pixel values represents peaks while the low (e.g. black) pixel values denotes valleys. The algorithm tries to grow the region areas by flooding the valleys and where different regions meet with each other are the watershed lines needed for image segmentation[49]. Watershed methods were applied to find the boundary between defect pixels and background pixels. We followed the official tutorial from OpenCV for performing the watershed and details of the approach can be found there[50]. We then fit the boundaries found from the Watershed algorithm to an ellipse. This fitting was done to match the approach used by the radiation defect analysis community, obtain a well-defined shape with simple geometric descriptors, and smooth out the otherwise rather rough boundaries found by the Watershed algorithm. The fitting was done with OpenCV's `fitEllipse()` function[51]. All codes were based with OpenCV[48] and by applying the second module we could get precise information about the defects' position, size, and orientations. The diameters and areas of defects are defined as follows, where *a* and *b* are



half the lengths of major and minor axes of the ellipse. The diameter of the a/2<111> and a<100> defects are defined as 2$a$. The diameter of the black dot is defined as twice the square root of ($ab$). The area of all defects is defined as $\pi ab$. The areal density is the sum of defect areas in a set of images divided by the total area of the set of images.

### *Data Availability*

We used a subset of published STEM images of the irradiated FeCrAl alloy system[5] (https://publish.globus.org/jspui/handle/ITEM/997) which were labeled and used in this study. The data are available at Figshare (https://doi.org/10.6084/m9.figshare.8266484) and the source code for the model is available on Github ( https://github.com/uw-cmg/multitype-defect-detection ). The dataset includes both the images and bounding boxes we used for this project. Data on Figshare also includes a CSV file with all data used in plots in this paper.

### *Supporting Information*

We showed the fitting results of all 12 testing images in section 1 of SI. In Section 2 of SI, we presented the labeling process that has been used in a previous study[5] and prepared a detailed instruction document to record our labeling process, which can be easily used for other defect images. In section 3 of SI, detailed statistics distribution of



the human labelling and machine predicting results of diameters and areal density were showed.

**Competing Interests:**

There are no competing interests in relation to the work described.

**Acknowledgments:**

We would like to thank Wisconsin Applied Computing Center (WACC) for providing access to CPU/GPU cluster, Euler. And special thanks to Colin Vanden Heuvel for helping us use GPUs and install software needed and sincere thanks to Vanessa Meschke for helping organize the undergraduate students participate in this research in summer 2018.


**Funding[1]:**

Research was sponsored by the Department of Energy (DOE) Office of Nuclear Energy, Advanced Fuel Campaign of the Nuclear Technology Research and Development program (formerly the Fuel Cycle R&D program). Neutron irradiation of FeCrAl alloys at Oak Ridge National Laboratory's High Flux Isotope Reactor user facility was sponsored by the Scientific User Facilities Division, Office of Basic Energy Sciences, DOE. Support for D. M. was provided by the National Science Foundation Cyberinfrastructure for Sustained Scientific Innovation (CSSI) program, award No. 1931298. Support for select undergraduate participants over some periods provided by the NSF University of Wisconsin-Madison Materials Research Science and Engineering Center (DMR 1720415) and the Schmidt Foundation.

---

[1] Notice: This manuscript has been authored by UT-Battelle, LLC, under contract DE-AC05-00OR22725 with the US Department of Energy (DOE). The US government retains and the publisher, by accepting the article for publication, acknowledges that the US government retains a nonexclusive, paid-up, irrevocable, worldwide license to publish or reproduce the published form of this manuscript, or allow others to do so, for US government purposes. DOE will provide public access to these results of federally sponsored research in accordance with the DOE Public Access Plan (http://energy.gov/downloads/doe-public-access-plan).



# Reference


1. Jenkins, M. . & Kirk, M. . *Characterisation of Radiation Damage by Transmission Electron Microscopy. Iop* **20002352**, (Taylor & Francis, 2000).

2. Jesse, S. *et al.* Big Data Analytics for Scanning Transmission Electron Microscopy Ptychography. *Sci. Rep.* **6**, 26348 (2016).

3. Kalinin, S. V., Sumpter, B. G. & Archibald, R. K. Big-deep-smart data in imaging for guiding materials design. *Nat. Mater.* **14**, 973–980 (2015).

4. Duval, L. *et al.* Image processing for materials characterization: Issues, challenges and opportunities. in *2014 IEEE International Conference on Image Processing, ICIP 2014* 4862–4866 (IEEE, 2014). doi:10.1109/ICIP.2014.7025985

5. Li, W., Field, K. G. & Morgan, D. Automated defect analysis in electron microscopic images. *npj Comput. Mater.* **4**, 1–9 (2018).

6. Park, C. & Ding, Y. Automating material image analysis for material discovery. *MRS Commun.* **9**, 545–555 (2019).

7. Groom, D. J. *et al.* Automatic segmentation of inorganic nanoparticles in BF TEM micrographs. *Ultramicroscopy* **194**, 25–34 (2018).

8. DeCost, B. L., Francis, T. & Holm, E. A. Exploring the microstructure manifold: Image texture representations applied to ultrahigh carbon steel microstructures. *Acta Mater.* **133**, 30–40 (2017).

9. Decost, B. L. & Holm, E. A. A computer vision approach for automated analysis





and classification of microstructural image data. *Comput. Mater. Sci.* **110**, 126–133 (2015).

10. DeCost, B. L., Jain, H., Rollett, A. D. & Holm, E. A. Computer Vision and Machine Learning for Autonomous Characterization of AM Powder Feedstocks. *Jom* **69**, 456–465 (2017).

11. DeCost, B. L. & Holm, E. A. Characterizing powder materials using keypoint-based computer vision methods. *Comput. Mater. Sci.* **126**, 438–445 (2017).

12. Marsh, B. P., Chada, N., Sanganna Gari, R. R., Sigdel, K. P. & King, G. M. The Hessian Blob Algorithm: Precise Particle Detection in Atomic Force Microscopy Imagery. *Sci. Rep.* **8**, 978 (2018).

13. Chowdhury, A., Kautz, E., Yener, B. & Lewis, D. Image driven machine learning methods for microstructure recognition. *Comput. Mater. Sci.* **123**, 176–187 (2016).

14. Vlcek, L., Maksov, A., Pan, M., Vasudevan, R. K. & Kalinin, S. V. Knowledge Extraction from Atomically Resolved Images. *ACS Nano* **11**, 10313–10320 (2017).

15. DeCost, B. L. & Holm, E. A. A large dataset of synthetic SEM images of powder materials and their ground truth 3D structures. *Data Br.* **9**, 727–731 (2016).

16. Gola, J. *et al.* Advanced microstructure classification by data mining methods. *Comput. Mater. Sci.* **148**, 324–335 (2018).

17. Liu, L. *et al.* Deep Learning for Generic Object Detection: A Survey. (2018).

18. Zou, Z., Shi, Z., Guo, Y. & Ye, J. Object Detection in 20 Years: A Survey. 1–39





(2019).

19.   Roberts, G. *et al.* Deep Learning for Semantic Segmentation of Defects in
      Advanced STEM Images of Steels. *Sci. Rep.* **9**, (2019).

20.   Anderson, C. M., Klein, J., Rajakumar, H., Judge, C. D. & B, L. K. Automated
      Classification of Helium Ingress in Irradiated X-750. 1–7 (2019).

21.   Rusanovsky, M. *et al.* Anomaly Detection using Novel Data Mining and Deep
      Learning Approach.

22.   Ahonen, T., Hadid, A. & Pietikäinen, M. Face description with local binary
      patterns: Application to face recognition. *IEEE Trans. Pattern Anal. Mach. Intell.*
      **28**, 2037–2041 (2006).

23.   Viola, P. & Jones, M. Rapid object detection using a boosted cascade of simple
      features. in *Proceedings of the 2001 IEEE Computer Society Conference on
      Computer Vision and Pattern Recognition. CVPR 2001* **1**, I-511-I–518 (IEEE
      Comput. Soc).

24.   Lecun, Y., Bengio, Y. & Hinton, G. Deep learning. *Nature* **521**, 436–444 (2015).

25.   Badrinarayanan, V., Kendall, A. & Cipolla, R. SegNet: A Deep Convolutional
      Encoder-Decoder Architecture for Image Segmentation. *IEEE Trans. Pattern
      Anal. Mach. Intell.* **39**, 2481–2495 (2017).

26.   Ziatdinov, M. *et al.* Deep Learning of Atomically Resolved Scanning
      Transmission Electron Microscopy Images: Chemical Identification and Tracking





Local Transformations. *ACS Nano* **11**, 12742–12752 (2017).

27.  Zafari, S., Eerola, T., Ferreira, P., Kälviäinen, H. & Bovik, A. Automated

Segmentation of Nanoparticles in BF TEM Images by U-Net Binarization and

Branch and Bound. in *Lecture Notes in Computer Science (including subseries*

*Lecture Notes in Artificial Intelligence and Lecture Notes in Bioinformatics)* **11678**

**LNCS**, 113–125 (Springer Verlag, 2019).

28.  Chen, D., Zhang, P., Liu, S., Chen, Y. & Zhao, W. Aluminum alloy

microstructural segmentation in micrograph with hierarchical parameter transfer

learning method. *J. Electron. Imaging* **28**, 1 (2019).

29.  Field, K. G., Briggs, S. A., Sridharan, K., Yamamoto, Y. & Howard, R. H.

Dislocation loop formation in model FeCrAl alloys after neutron irradiation below

1 dpa. *J. Nucl. Mater.* **495**, 20–26 (2017).

30.  Parish, C. M., Field, K. G., Certain, A. G. & Wharry, J. P. Application of STEM

characterization for investigating radiation effects in BCC Fe-based alloys. *J.*

*Mater. Res.* **30**, 1275–1289 (2015).

31.  Field, K. G., Hu, X., Littrell, K. C., Yamamoto, Y. & Snead, L. L. Radiation

tolerance of neutron-irradiated model Fe-Cr-Al alloys. *J. Nucl. Mater.* **465**, 746–

755 (2015).

32.  Zinkle, S. J. & Busby, J. T. Structural materials for fission & fusion energy.

*Materials Today* **12**, 12–19 (2009).





33. Yao, B., Edwards, D. J. & Kurtz, R. J. TEM characterization of dislocation loops in irradiated bcc Fe-based steels. *J. Nucl. Mater.* **434**, 402–410 (2013).

34. Ren, S., He, K., Girshick, R. & Sun, J. Faster R-CNN: Towards Real-Time Object Detection with Region Proposal Networks. *IEEE Trans. Pattern Anal. Mach. Intell.* **39**, 1137–1149 (2015).

35. Zhao, Z.-Q., Zheng, P., Xu, S.-T. & Wu, X. Object Detection With Deep Learning: A Review. *IEEE Trans. Neural Networks Learn. Syst.* 1–21 (2019). doi:10.1109/TNNLS.2018.2876865

36. Dean, J. *et al.* Large Scale Distributed Deep Networks. in *Advances in Neural Information Processing Systems 25* (eds. Pereira, F., Burges, C. J. C., Bottou, L. & Weinberger, K. Q.) 1223–1231 (Curran Associates, Inc., 2012).

37. Jia, Y. *et al.* Caffe. in *Proceedings of the ACM International Conference on Multimedia - MM '14* 675–678 (ACM Press, 2014). doi:10.1145/2647868.2654889

38. Cheng, Y., Wang, D., Zhou, P. & Zhang, T. A Survey of Model Compression and Acceleration for Deep Neural Networks. (2017).

39. Chen, P.-H. C. *et al.* An augmented reality microscope with real-time artificial intelligence integration for cancer diagnosis. *Nat. Med.* 1–5 (2019). doi:10.1038/s41591-019-0539-7

40. Pan, S. J. & Yang, Q. A Survey on Transfer Learning. *IEEE Trans. Knowl. Data Eng.* **22**, 1345–1359 (2010).





41. Field, K. G. *et al.* Heterogeneous dislocation loop formation near grain boundaries in a neutron-irradiated commercial FeCrAl alloy. *J. Nucl. Mater.* **483**, 54–61 (2017).

42. Schindelin, J. *et al.* Fiji: An open-source platform for biological-image analysis. *Nat. Methods* **9**, 676–682 (2012).

43. Schneider, C. A., Rasband, W. S. & Eliceiri, K. W. NIH Image to ImageJ: 25 years of image analysis. *Nat. Methods* **9**, 671–675 (2012).

44. Niitani, Y., Ogawa, T., Saito, S. & Saito, M. ChainerCV: a Library for Deep Learning in Computer Vision. 1217–1220 (2017). doi:10.1145/3123266.3129395

45. Module: exposure — skimage v0.18.0 docs. Available at: https://scikit-image.org/docs/stable/api/skimage.exposure.html#skimage.exposure.equalize_adapthist. (Accessed: 23rd April 2021)

46. Module: filters — skimage v0.19.0.dev0 docs. Available at: https://scikit-image.org/docs/dev/api/skimage.filters.html#skimage.filters.gaussian. (Accessed: 23rd April 2021)

47. Shorten, C. & Khoshgoftaar, T. M. A survey on Image Data Augmentation for Deep Learning. *J. Big Data* **6**, 60 (2019).

48. Garrido, G. & Joshi, P. *OpenCV 3.x with Python By Example Second Edition Make the most of OpenCV and Python to build applications for object recognition and augmented reality.* (2018).





49.    Kornilov, A. & Safonov, I. An Overview of Watershed Algorithm

        Implementations in Open Source Libraries. *J. Imaging* **4**, 123 (2018).

50.    OpenCV: Image Segmentation with Watershed Algorithm. Available at:

        https://docs.opencv.org/master/d3/db4/tutorial_py_watershed.html. (Accessed:

        24th March 2021)

51.    opencv/fitellipse.py at master · kipr/opencv. Available at:

        https://github.com/kipr/opencv/blob/master/samples/python/fitellipse.py.

        (Accessed: 21st April 2021)